\documentclass{article} 
\usepackage{mefomo_2023,times}

\usepackage{amsmath,amsfonts,bm}

\def\eqref#1{equation~\ref{#1}}

\def\floor#1{\lfloor #1 \rfloor}
\def\1{\bm{1}}

\DeclareMathAlphabet{\mathsfit}{\encodingdefault}{\sfdefault}{m}{sl}
\SetMathAlphabet{\mathsfit}{bold}{\encodingdefault}{\sfdefault}{bx}{n}

\usepackage{hyperref}
\usepackage{url}
\usepackage{times}
\usepackage{latexsym}
\usepackage{graphicx}
\usepackage{amsfonts}
\usepackage{amsmath}
\usepackage{amssymb}
\usepackage{booktabs} 
\usepackage{multirow}
\usepackage{cleveref}
\usepackage{xspace}
\usepackage{enumitem}
\usepackage{mathtools}
\usepackage{array}

\crefformat{section}{\S#2#1#3}
\crefmultiformat{section}{\S\S#2#1#3}{ and~#2#1#3}{, #2#1#3}{, and~#2#1#3}
\long\def\ignore#1{}

\newcommand{\model}{\textsc{DENSE}\xspace}

\newcommand{\cutsectiondown}{\vspace*{-0.0in}}

\newcommand{\cutsubsectionup}{\vspace*{-0.0in}}
\newcommand{\cutsubsectiondown}{\vspace*{-0.0in}}

\newcommand{\cutparagraphdown}{\vspace*{-0.15in}}

\newcommand{\cutcaptionup}{\vspace*{-0.07in}}
\newcommand{\cutcaptiondown}{\vspace*{-0.10in}}

\title{Exploring Demonstration Ensembling for In-context Learning}

\author{Muhammad Khalifa\thanks{$\,\,\,$Correspondence to \tt{khalifam@umich.edu}}\hspace{3pt}, Lajanugen Logeswaran$^\dagger$, \textbf{Moontae Lee}$^{\dagger\ddagger}$, \\ 
\textbf{Honglak Lee}$^{*\dagger}$\textbf{,} \textbf{Lu Wang}$^*$ \\
University of Michigan$^*$, LG AI Research$^\dagger$, University of Illinois at Chicago$^\ddagger$
}

\iclrfinalcopy 

\begin{document}
\maketitle
\begin{abstract}
In-context learning (ICL) operates by showing language models (LMs) examples of input-output pairs for a given task, \textit{i.e.,} demonstrations. The standard approach for ICL is to prompt the LM with concatenated demonstrations followed by the test input. This approach suffers from some issues. First, concatenation offers almost no control over the contribution of each demo to the model prediction. This can be sub-optimal when some demonstrations are irrelevant to the test example. Second, due to the input length limit of some transformer models, it might be infeasible to fit many examples into the context, especially when dealing with long-input tasks. In this work, we explore \textbf{D}emonstration \textbf{Ense}mbling (\model) as an alternative to simple concatenation. \model predicts outputs using subsets (\textit{i.e.}, buckets) of the demonstrations and then combines the output probabilities resulting from each subset to produce the final prediction. We study different ensembling methods using GPT-j \citep{wang2021gpt} and experiment on 12 language tasks. Our experiments show weighted max ensembling to outperform vanilla concatenation by as large as 2.4 average points.\footnote{Code available at \url{https://github.com/mukhal/icl-ensembling}.} 
\end{abstract}

\section{Introduction}
\cutsectiondown

Large-scale language model (LM) pre-training on large corpora is currently dominating the NLP scene. One impressive aspect of such large language models (LLMs) is their capability to do in-context learning \citep{brown2020language} by conditioning the model on a few examples (\textit{i.e.,} demonstrations) of the desired task and then asking the LM to predict the label for a given input. 

The standard approach for feeding in-context demonstrations (demos, \textit{for short}) to the LM is by \textit{concatenating} the task examples \citep{brown2020language,min2022rethinking,fant22}. While simple, concatenation suffers from a few drawbacks. First, it provides no control over each demo's contribution to the model's output, which is left to the attention weights to decide. Second, the concatenation of demos can easily use up the context window of transformer-based models, especially when we have access to many demonstrations or when dealing with lengthy inputs. Lastly, it has been shown that LLMs are sensitive to the ordering of the demonstrations \cite{calibrate2021,fant22}, and a long chain of concatenated demos can indeed exacerbate this problem.

In this work, we explore an alternative to the concatenation approach, which is to prompt the model with demonstrations in an ensembling approach. In particular, we partition the examples into non-empty subsets or buckets and then combine the predictions obtained from each bucket to obtain the final prediction. We investigate three different ensembling methods to combine the predictions from different buckets including a clustering-based approach to partition the examples. Experiments on 7 different language tasks show that ensembling can outperform the standard concatenation approach.

\cutsectiondown

\section{Related Work}
\cutsectiondown
This work is related to work that aims to improve few-shot learning with LLMs \citep{metaicl,retrieveicl22,fant22}. For instance, \citet{perez2021true} try to find optimal prompts using techniques such as cross-validation and minimum description length. \citet{channel22} applied demonstration ensembling for text classification in a limited setting. This paper, on the other hand, explores the more generalized ensembling setting with different bucket sizes and different types of tasks. 
\citet{wang2022rationale} explore rationale-augmented ensembles, where different in-context demonstrations are augmented with LM-generated rationales. Different from our work, the ensembling is done over the rationales in the examples, while we ensemble the examples themselves. \citet{qin2021} trained mixtures of soft prompts for knowledge extraction from language models. 
 Our prior work~\citep{khalifa2022lepus} also explored demonstration ensembling to in-context learn to rerank document paths for multi-hop question answering. 
\cutsectiondown

\section{Demonstration Ensembling}
\cutsectiondown

\begin{figure}[t!]
    \centering
    \includegraphics[width=13.5cm]{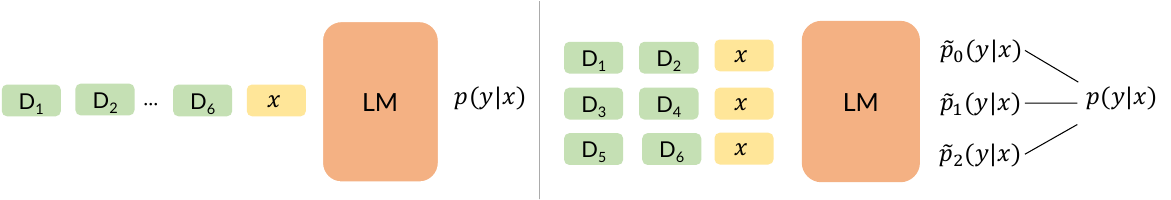}
    \cutcaptionup
    \caption{In-context learning with six demonstrations.  \textbf{Left:} The standard concat-based approach for feeding the examples \citep{brown2020language}. \textbf{Right:} Ensembling with three buckets of size two each. For a given label $y$, the probability $\tilde{p}_i(y|x)$ is computed using the $i$-th bucket. All probabilities are ensembled to give the final probability $p(y|x)$.
    }
    \cutcaptiondown
    \label{fig:my_label}
\end{figure}

We assume a list of $n$ demonstrations $\mathcal{D} = \langle(x_1, y_1), .., (x_n, y_n)\rangle$, where $x_i$ and $y_i$ are the demonstration input and ground-truth output or label, respectively. We now formalize our approach for demonstration ensembling.
\subsection{Bucket Allocation}
\label{sec:bucket-alloc}

\model allocates the $n$ demos in $\mathcal{D}$ to $b$ non-empty buckets $\{\mathcal{B}_0, \mathcal{B}_1, ..., \mathcal{B}_{b-1}\}$. More precisely, if each bucket has $\gamma$ demos, then $\mathcal{B}_i$ is assigned the demos $\mathcal{D}_{\gamma i:\gamma(i+1)-1}$. We predict a set of probabilities of a label $y$ by \textit{separately} conditioning the LM on the different buckets along with the test input $x$. Formally, for bucket $\mathcal{B}_i$, we predict $\tilde{p}_{i}(y | x)$ as:
\begin{equation*}
\begin{split}
    \tilde{p}_{i}(y | x) &= P_{LM}(y | \mathcal{B}_i, x)\\
    \end{split}
\end{equation*}

\noindent
 The aggregate probability of the label $y$ is proportional to the output of an ensembling operator $\Phi$ that combines different bucket probabilities: 

\begin{equation}
    P(y | x) \propto \Phi(y | \tilde{p}_{0}(y | x), \hdots, \tilde{p}_{B-1}(y | x), x)
\end{equation}

\noindent
Where $\Phi$ is a function that takes in the probabilities $\tilde{p}_{0}(y | x), \hdots, \tilde{p}_{B-1}(y | x)$ and the test example $x$, and computes a (possibly unnormalized) probability of the output label $y$. For brevity, we will just use $\Phi(y | x)$ from now on.

\subsection{Ensembling Method}
\cutsubsectiondown
We assume each bucket $\mathcal{B}_i$ has a normalized importance weight $w_i$ assigned to it where $\sum_{i=0}^{b} w_i = 1$. One form of $\Phi$ is the product operator in which $P(y|x)$ corresponds to a \textit{product-of-experts} \citep{hinton2002training}:
\begin{equation}
    \label{eq_poe}
    \Phi^{\operatorname{PoE}}(y | x) = \prod_{i=0}^{b} \tilde{p}_{i}(y | x)^{w_i}
\end{equation}

\noindent 
In addition, we can explore a mixture-of-experts formulation:
\begin{equation}
\label{eq_moe}
     \Phi^{\operatorname{MoE}}(y | x) = \sum_{i=0}^{b} w_i \tilde{p}_{i}(y | x)
\end{equation}
We also explore max ensembling, which uses the \textit{most confident} prediction probability across different buckets: 
\begin{equation}
    \label{eq_max}
    \Phi^{\operatorname{max}}(y | x) = \max_j{w_j \tilde{p}_{j}(y | x)}
\end{equation}
\noindent

\subsection{Bucket Weighting}
\cutsubsectiondown
Inspired by recent work \citep{gao-etal-2021-making,whatmakes2022} that has shown that demonstrations that are more similar to the input perform better than distant ones, 
we weigh each bucket using the average of the similarity of its examples with the input $x$: 
\begin{equation}
    \label{eq:sim_weight}
    w_i = \frac{1}{|\mathcal{B}_i|} \sum_{(x_j, y_j) \in \mathcal{B}_i} \text{cos}(x_j^e, x^e)
\end{equation}

\noindent where $\text{cos}$ is the cosine similarity and $x^e$ is the embedding of the $x$.

\subsection{Clustering Demonstrations}
\cutsubsectiondown
\label{sec:clustering}
While the bucket construction approach explained in \Cref{sec:bucket-alloc} constructs buckets arbitrarily based on the order of the demos in $\mathcal{D}$, one heuristic is to use similarity information between demos to construct buckets. We experiment with k-means clustering \citep{hartigan1979algorithm} to construct buckets. More precisely, we apply k-means over vector representations of the demonstrations to obtain $b$ clusters and then use each cluster as a bucket.\footnote{Note that in this case, not all buckets will have the same number of demos.} Each bucket can operate as a semantically coherent expert. We refer to this approach as \textbf{similar-together} bucket allocation. 

\begin{figure}[t!]
    \centering
    \includegraphics[width=13.7cm]{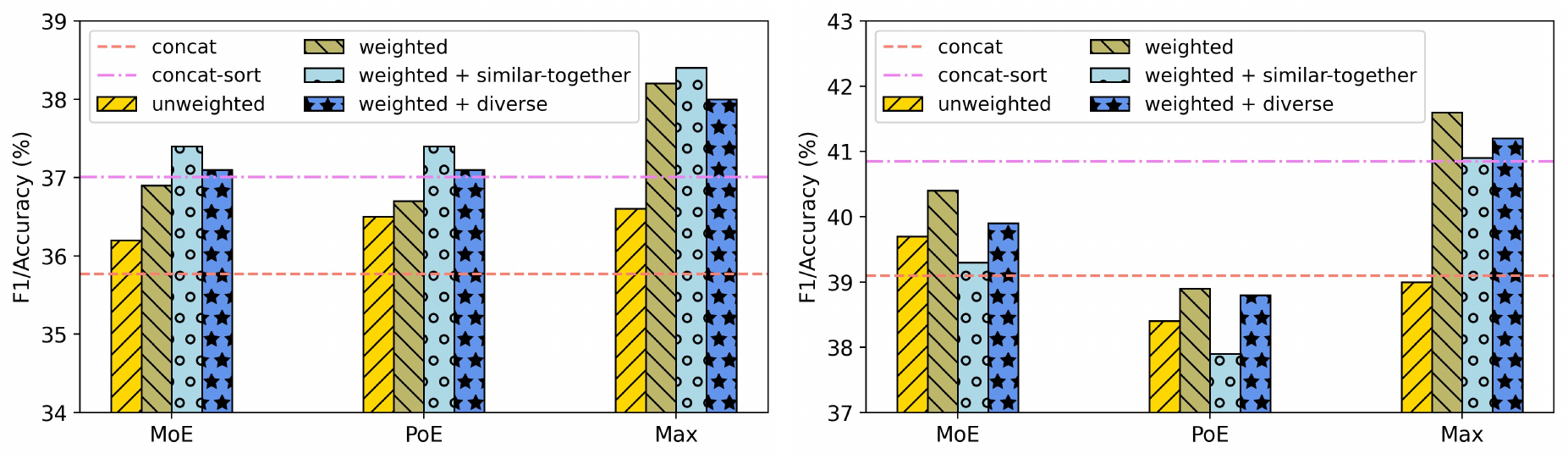}
    \cutcaptionup
    \caption{6-shot \textbf{(Left)} and 10-shot \textbf{(Right)} performance of different ensembling methods and concatenation. Metrics are averaged over three seeds of demos, 12 datasets, and different numbers of buckets. \textbf{(Un)weighted} indicates whether we use similarity with the input examples to weigh the contribution of each ensemble. \textbf{Similar-together} bins and \textbf{Diverse} buckets are achieved through k-means clustering as explained in ~\ref{sec:clustering}}
    \cutcaptiondown
    
    \label{fig:main-results}
\end{figure}

As opposed to maximizing the similarity between the demos within a given bucket, instead, we can maximize dissimilarity to achieve diverse buckets. To do that, we use K-means to cluster demos into $\floor{n/b}$ clusters, each with $b$ demos.\footnote{We assume $n$ is always divisible by $b$ for simplicity.} Then, we construct $b$ buckets by picking a unique demo from each cluster.\footnote{Here, we use a constrained version of k-means \citep{bradley2000constrained} to make sure we get exactly $b$ demos in each k-means cluster.} Having diverse buckets might result in a prediction that is less biased towards a certain category of demonstrations. We refer to this approach as \textbf{diverse} bucket allocation. 

Besides yielding better bucket allocation, clustering makes bucket assignment less sensitive with respect to the demonstration order in $\mathcal{D}$. As a result, it can greatly reduce the sensitivity to the order of the demos studied in previous work \citep{calibrate2021,fant22}.
\cutsubsectiondown

\label{sec:combining-prob}

\section{Experiments and Results}
\cutsectiondown

\subsection{Experimental Setup}
\cutsubsectiondown

\paragraph{Data.} We experiment with 12 datasets in total. Details on the datasets, metrics used, and the number of evaluation examples can be found in \Cref{app:datasets}.
\paragraph{Baselines.} We compare ensembling to two concatenation-based settings. \textbf{Concat} is the standard approach for ICL, which concatenates the demonstrations in an arbitrary order, and \textbf{Concat-sort} which sorts the demos based on similarity with the input (i.e., more similar demos come later). Concat-sort uses similarity with the input, allowing for a fair comparison with the weighted ensembling methods.

\cutparagraphdown
\paragraph{Model.} For all the experiments, we use GPT-j (6B) \citep{wang2021gpt}. To compute embeddings of examples for similarity calculations, we use a fine-tuned 6-layer MiniLM \citep{wang2020minilm}.\footnote{\url{https://huggingface.co/sentence-transformers/all-MiniLM-L6-v2}.} Our experimental setup is detailed in \Cref{app:exp-setup}.
\cutparagraphdown
\paragraph{Number of demonstrations and bucket count.} We experiment with number of examples $n=6,10$. For $n=6$, we use ensembling with bucket counts $b=2,3,6$, and for $n=10$, we set  $b=2,5,10$. We note that the concat method in \cite{brown2020language} is a special case of ensembling with $b=1$.
\cutsubsectiondown

\begin{figure}[t!]
    \centering
    \includegraphics[width=14.0cm]{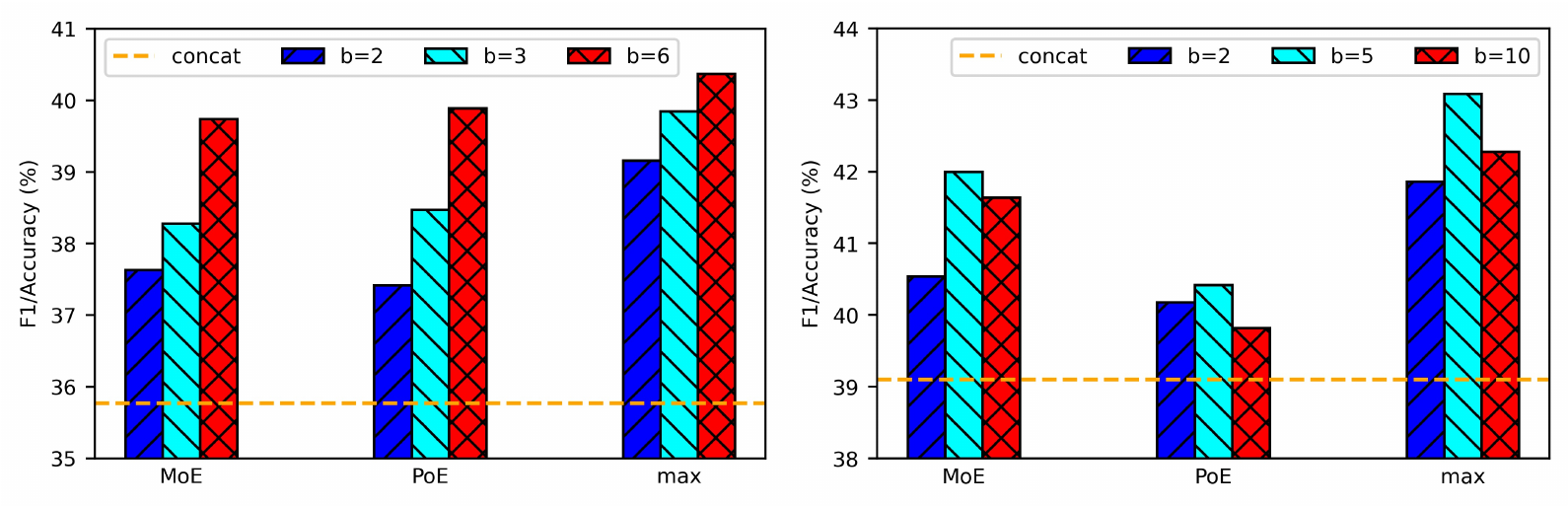}
    \cutcaptionup
    \caption{6-shot \textbf{(Left)} and 10-shot \textbf{(Right)} performance with different bucket count $b$. We show performance with weighted MoE, PoE, and Max ensembling. Performance is averaged over 7 tasks and over 3 different seeds of demos.
    }
    \cutcaptiondown
    \label{fig:nbuckets}
\end{figure}

\subsection{Comparing Ensembling Methods}
\cutsubsectionup
\Cref{fig:main-results} compares the performance of the concat approach against various ensembling methods and in the 6- and 10-shot cases. We observe that \textbf{unweighted} ({i.e.} $w_i = \frac{1}{b}$ for all $i$), PoE, MoE, and max ensembling outperform the concat baseline. Precisely, max ensembling outperforms concat by 0.8 average points in the 6-shot setting while MoE outperforms it by 0.6 in the 10-shot setting. However, unweighted ensembling underperforms the concat-sort baseline.

Furthermore, \textbf{weighing} the buckets based on the similarity with inputs boosts the ensembling performance in \textit{all} cases. In the 6-shot case, weighted max ensembling outperforms concat and concat-sort by 2.4 and 1.2, respectively. In the 10-shot setting, weighted max ensembling outperforms concat and concat-sort by 1.9 and 0.75 respectively.  

Lastly, we study the effect of bucket allocation based on \textbf{clustering} the demonstrations, where similar-together clustering gives a consistent boost to weighted ensembling methods in the 6-shot case. However, we do not have conclusive evidence as to which clustering strategy is best. We find the performance of the clustering strategy to vary depending on the task and the total number of demonstrations. For instance, \textbf{diverse} always outperforms \textbf{similar-together} allocation in the 10-shot case, which is not the case for the 6-shot setting. This is likely because having more demos allows for more diverse buckets. We leave it to future work to explore different methods of bucket allocation. \Cref{fig:improvement-per-task} in \Cref{app:detailed results} shows per-task improvement obtained by ensembling.

\cutsubsectiondown
\subsection{Buckets Count}
\cutsubsectiondown
Here we study what role the bucket count $b$ plays in the performance of ensembling. \Cref{fig:nbuckets} shows the effect of changing the bucket count on the performance. Using a small $b$ seems to perform worse across the board. Interestingly, the performance improves as $b$ increases for all ensembling methods in the 6-shot setting. 

%
\cutsubsectiondown


\cutsubsectiondown

\section{Conclusion}
In this work, we explore an alternative to the popular in-context learning paradigm where examples are concatenated and provided to a language model. We show through experiments on 12 language tasks that ensembling, where examples are partitioned into buckets and a final prediction is made by combining predictions from each bucket, yields better performance over concatenation. In particular, we find that max ensembling performs best compared to product-of-experts and mixture-of-experts. 
In addition, we analyze the effect of varying different aspects of ensembling such as the number of buckets and bucket construction strategies.

\bibliography{iclr2023_conference}
\bibliographystyle{iclr2023_conference}

\newpage
\appendix

\section{Datasets}
\label{app:datasets}
\begin{table*}[h!]
    \centering
    \small
    \begin{tabular}{cccc}
    \toprule
    \textbf{Dataset} & \textbf{Task} & \textbf{Metric} &  \textbf{\# Eval} \\
    \midrule
    Glue-SST2 \citep{sst2} &  sentiment analysis & macro F$_1$& 872 \\
    Medical questions pairs \citep{medicalpairs} &  paraphrase detection & macro F1& 610 \\
    Glue-MRPC \cite{levesque2012winograd} & paraphrase detection & macro F1& 408 \\
    Climate Fever \citep{climatefever} & fact verification & macro F1& 307 \\
    SICK \citep{sick}  & NLI & macro F1& 495 \\
    Glue-WNLI \cite{dolan2005automatically} & NLI & macro F1& 71 \\
    Hate speech18 \cite{hatespeech18} & hate speech detection & macro F1& 2141 \\
    TweetEval-stance (feminism) \citep{tweeteval} & stance detection & macro F1& 67 \\ 
    OpenbookQA \cite{openbookqa} & question answering & accuracy & 500 \\
    ARC \cite{clark2018think} & question answering & accuracy & 299 \\
    QUAREL \cite{tafjord2019quarel} & question answering
       & accuracy & 278 \\

    CODAH \cite{chen-etal-2019-codah} & sentence completion & accuracy & 556 \\

    \bottomrule
    \end{tabular}
    \caption{Datasets, tasks, metrics, and the number of evaluation examples for each dataset. }
    \label{tab:my_label}
\end{table*}

\section{Experimental Setup}
\label{app:exp-setup}
We run few-shot inference using fp16 half-precision. All experiments are run on a workstation with 4 Nvidia A100 GPUs with a batch size of 16. We use the GPT-j checkpoint provided by Huggingface.\footnote{\url{https://huggingface.co/EleutherAI/gpt-j-6B/tree/float16}} For clustering, we use the K-means implementation provided by sklearn.\footnote{\url{https://scikit-learn.org/stable/modules/generated/sklearn.cluster.KMeans.html}} For constrained K-means, we use this implementation.\footnote{\url{https://github.com/joshlk/k-means-constrained}}

\section{Detailed Results}
\label{app:detailed results}
\begin{figure}[ht!]
    \centering
    \includegraphics[width=12.1cm]{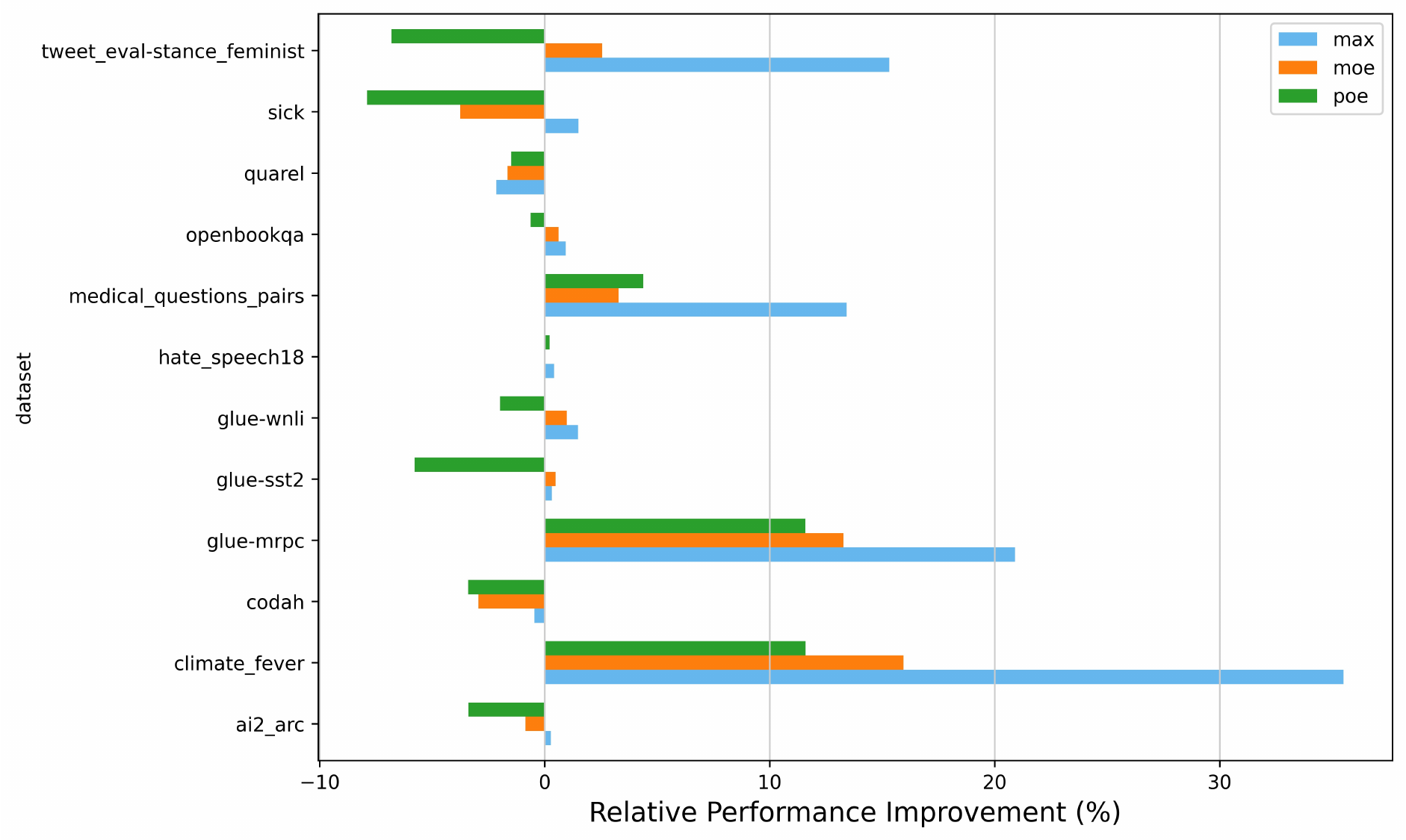}
    \cutcaptionup
    \caption{Relative performance improvement resulting from different ensembling methods shown per task. The improvement is aggregated over a different number of examples $6,10$, different numbers of buckets, and different seeds.}
    \cutcaptiondown
    \label{fig:improvement-per-task}
\end{figure}

\Cref{fig:improvement-per-task} shows relative improvement obtained by different \textbf{weighted} ensembling approaches over the concatenation approach.

\end{document}